# Has China caught up to the US in AI research? An exploration of mimetic isomorphism as a model for late industrializers

Chao Min
Nanjing University, Nanjing, China
0000-0002-0627-995X

Yi Zhao
Nanjing University of Science and Technology, China
0000-0003-1050-7051

Yi Bu
Peking University, China

Ying Ding
University of Texas at Austin, USA
0000-0003-2567-2009

Caroline S. Wagner*
The Ohio State University, Columbus, OH USA
0000-0002-1724-8489

(Chao Min and Yi Zhao contributed equally to this study)

Keywords: artificial intelligence; AI; computer science; government investment; research and development; national capacity; competition

*corresponding author wagner.911@osu.edu

# Has China caught up to the US in AI research? An exploration of mimetic isomorphism as a model for late industrializers

Abstract

Artificial intelligence represents a 21st century critical technology. We test for and find rapid learning and differentiation characterizing China's AI development process as a model with a better fit than the Foreign Direct Investment/export-oriented growth of earlier Asian industrializers. China outproduces the USA in numbers of AI papers, but quality measures show China lagging the US; even so, China's AI development is impressive. China's extremely rapid development of AI has benefitted from the worldwide movement towards open access to algorithms and papers, China's extensive diaspora and returnees, and weakly protected data. Applying a measure of China's imitation of US research, we find the time lag between China and the USA in AI research topics disappeared in 2018 suggesting that China has closed an important gap with the USA and may now be on an independent trajectory. This paper compares China (only) against USA (only), although we note that collaborative work between these two nations is more highly cited than work by either one separately.

# Introduction

The rapid rise of China provides an opportunity to gain insight into a nation developing its science and technology capacity within an established global science system. It is well known that nations invest in science and technology for prestige, to aid economic growth, improve health and infrastructure, and strengthen national security. Riberio et al. (2010) followed a long line of analysis (Freeman, 1988; Nelson, 1983; Lundvall, 1988) showing that science and technology fuel economic growth. Japan and South Korea followed a strategy of foreign-direct investment, export-led growth, and import substitution to grow their S&T systems. China offers the opportunity to study technology development in the "catching up" stage.

Malerba and Nelson (2011) posited that those nations seeking to catch up in science and technology engage in a learning process; they expected that significant differences in variables and mechanisms would manifest across sectors in a democratic/market-based political economy. China may exhibit significant differences in AI growth based upon its latecomer status and centralized economy model. This paper examines China's efforts to "catch up" in artificial intelligence (AI) over the past 20 years by comparing them to those of the United States, which has been the world leader in AI. China outlined a comprehensive policy supporting artificial intelligence (AI) research and development that commits to being a global leader in the field by 2025, becoming a world leader in AI innovation by 2030, and using AI to improve the country's overall competitiveness by 2045.

To support its AI goals, the Chinese government has established a number of national-level AI research institutes and innovation centers and has launched well-funded initiatives to support the development of AI technologies and applications (Hannas & Chang, 2023). AI policy states a commitment to responsible and ethical AI, with a focus on protecting privacy, security, and social stability. The plan includes provisions for international collaboration, including cooperation with international organizations and participation in international AI conferences and forums.

According to Amsden (1989), late industrializers who want to expand their science and technology bases invest in research and development in order to imitate and catch up to the leaders. In a study of South Korea, Amsden found that firms developed links with foreign counterparts to manufacture, first lower-value, then higher-value-added products in a process of incremental improvement. An export-oriented strategy brought in capital and augmented learning. Hikino and Amsden (1994) suggested that this process constituted a common method among late industrializers of borrowing technology from more advanced nations, often through foreign direct investment and associated technology transfer or spillover.

This raises the question of the process of late technological development in China and whether it follows the model suggested by Amsden (1992) and Hikino and Amsden (1994). At first glance, China shares the characteristics of late industrializers: making educational investments, engaging in export-led growth, and stepwise incremental technological change through Foreign Direct Investment (FDI). However, China[i] has notable differences from the model used by South Korea and other Asian Tigers. Within China, the method of seeking FDI, as South Korea did, was not viewed as the most effective means of technology development—at least not as the sole method (Zhou, 2010). Within China, policymakers are committed to indigenous development and productivity enhancement through education and investment, supported by a process of global engagement of students and scholars with collaborations occurring across borders and within China, suggesting that internal development may play a greater role in China than was the case for other Asian industrializers. The differences cause us to question whether the late-industrialization theory of political economy, developed to explain the long growth processes of the West and the more rapid growth of East Asian capitalist nations, may not apply to China.

Moreover, China's development of a political economy differs from South Korea in another way. Where Amsden suggested that businesses led Korea's development process, the Chinese Communist Party (CCP) and government ministries have appeared to lead China's development, with business as a secondary player. China's economy is unlike Korea's (or other Asian Tigers) in that the Chinese state continues to be the "ultimate economic decision-making authority…" (Zheng & Huang, 2018, p. 4) where Korea (and earlier, Japan) was business-led and market-based with companies receiving aid from government.

[i] Here we mean mainland China. Similarly hereafter.

We propose a process of rapid imitation and differentiation may better reflect China's development than does the FDI/export-oriented growth. Drawing from organization theory, we suggest that mimetic isomorphism, a concept proposed by DiMaggio and Powell (1983), can be expanded to apply to late industrializer's 'catching up' phase. The original concept applied isomorphism at the institutional level, particularly within industries. We propose to expand the concept to policy and policy implementation at the national level. We develop and apply tests to assess whether China's process can be shown to follow an isometric pattern, following upon stated political intent to catch up with the United States in AI. We expect to show that isomorphism better fits China's actions than does the FDI-led model derived from the Korean experience.

The catching up process involves multi-sectoral investments in developing financial systems, education, public sector research and government programs (Nelson 2006), all of which China has undertaken in the past two decades. In addition to many reforms, since 2016, Chinese public policy has targeted Artificial Intelligence (AI) for development as one of the cutting-edge technologies in the 21st century. Chinese industry started later than other nations to develop this capacity, but catch-up has been rapid (Ernst, 2020). For any nation, AI is projected to have a transformative impact on social and economic developments in many fields such as manufacturing, transportation, health, and education. More than 60 countries have issued AI policies, according to the Organization for Economic Co-operation and Development (OECD) Artificial Intelligence Observatory.

China has built significant science and technology capacity over the past three decades to join world nations in targeting this and other technologies for development.

China began this journey with information technology investments (very well summarized by Gu (1997)) and more recently with specific initiatives (described by Roberts et al., 2019). Keller (2004) showed that foreign sources account for most of the world's creation of new technologies in catch-up countries. Keller (2004) further suggested that no 'global pool of technology' exists; development remains both local and highly tacit. Domestic investment is necessary. In 'catching up' countries, foreign direct investment has been seen as the mechanism of diffusion (Fu et al., 2011), although, for China, this needs further testing. If this is the case, it suggests that China would look to the United States, the world leader, for technology to grow its AI sector.

Artificial Intelligence became an explicit focus of Chinese science policy in 2018 with the publication of the "Made in China 2025" plan to which the government committed $300 billion. Naming AI as a focal area of technological innovation emerges from years of related national plans, such as the 863 Plan (Cao et al., 2019), which contained specific targets for advances in information technologies, electronics developments, and computing. Early plans exhibited mixed success in practice, but more recent plans have been more coherent, more representative of multi-sectoral consensus, and well-funded. In 2013, the Strategic Emerging Industries Initiative was issued, shifting China from a techno-nationalist orientation towards more diverse and economically tuned priorities (USCC, 2016). Among the priorities was a focus on "new information technology industry" which, while not naming artificial intelligence directly, instructs the Ministry of Industry and Information Technology (MIIT) to accelerate the construction of next-generation information technologies (MIIT, 2012).

# China and the U.S. in the Global AI research

Previous studies have used bibliographic datasets to understand the patterns of global AI research and its evolution. In terms of productivity, Tang et al. (2020) painted a nuanced picture on the speed of producing AI innovations. Specifically, their quantitative analyses estimated that three AI papers were submitted to arXiv per hour, and ~5.3 new scientists started to work on AI each hour in 2019. Yuan et al. (2020a) studied the evolution of AI research from three different, yet interrelated, perspectives, namely trend, mobility, and scientific collaboration. They found that, in a global view, people have been shifting from theoretical to applied sides of AI in their research and that elite AI scientists have been highly clustered in the coauthorship network. In another work, taking a different approach, Rahkovsky, Toney, Boyack, Klavans, and Murdick (2021) examined research funding of AI by investigating six large funding organizations among the U.S., European Union, China, and Japan. These large-scale analyses untangled some regional differences for AI research: For example, National Science Foundation of the U.S. (NSF) and European Research Council (ERC) tend, per their mission, to support fundamental advancement of AI and machine learning instead of their applications. Shao et al. (2020) illustrated company-level competition in the AI domain, suggesting that IBM is losing its academic influence advantage than before; they also found increased asymmetry of institutional attractiveness.

The rapid development of AI also co-evolves with other related disciplines. For instance, Frank et al. (2019) concluded that recent AI studies tended to reference mathematics and computer science instead of philosophy, geography, and art (which they historically cited); these indicate a distinct flow of knowledge in a more technical direction, as well as different patterns of the source and diffusion of AI innovations.

A short essay in Science (Brainard & Normile, 2022) reported on scholarship (Wagner, Zhang, & Leydesdorff, 2022) that China has risen to first place in most cited scientific publications. This milestone for most cited papers is indicator of China's science development, following China's rise to the top in number of publications (Wagner & Cai, 2022). In many disciplines, the U.S. and China show intense competition in frontiers of science, technology, and engineering (Cady & Etzioni, 2019). Bibliometric analyses have revealed that China published more AI papers and secured more AI patents than U.S. (Li, Tong, & Xiao, 2021).

China's AI publications and patents have not risen to US levels of citation (e.g., Simonite, 2019; Thomas & Murdick, 2020). Given its shorted engagement in the field, one would not expect this achievement. Quantitative analyses shows that the average number of citations of China's AI publications is lower than that of U.S. with field and age normalization, but that gap is decreasing continuously (O'Meara, 2019). Regarding international citations, the U.S. still had advantages, compared with China (Acharya & Dunn, 2022). A data-driven prediction made in 2021 said that China would become the leader in the top 1% most cited AI scientific publications in 2023 (Yang & Etzioni, 2021). In terms of scientist mobility, based on the ArnetMiner dataset recording the bibliographic information of AI researchers, Yuan et al. (2020a) demonstrated that scientists tend to move out from China more than to move in, compared with the in-out statistics of the U.S.

Yet, there are only a limited number of related studies that discussed potential mechanisms explaining these quantitative measurements on China vs. U.S. AI research. In a recent study, Lundvall and Rikap (2022) argued that China's catching-up in the domain of AI is partly interpreted as its corporate and national innovation system (e.g., "Made in China 2025" policy in China, as mentioned in Li et al. (2021)); this is particularly reflected by the comparisons between the two tech giants in China, namely Alibaba and Tencent, and their U.S. counterparts.

# Methodology and Data

Analysts often compare cross-country inputs (research and development spending and trained people) and outputs (articles, citations, and patents) measures. While the field of AI is highly internationalized, national borders continue to represent policy and legal differences as well as cultural and natural assets that influence successful progress that provide meaningful points of comparison. To enable a comparison, this study isolates indicators for China and the United States to test for imitation. Collaborative work between them will be addressed in a separate article.

The key aspects of national power in AI are expected to look different from other types of technology (Horowitz et al., 2018): AI is an enabler, or a platform technology (Tassey, 2008), and a disruptive force with diverse capabilities for applications in industry and the military. "AI will augment the national power of those countries that are able to identify, acquire, and apply large datasets of high economic and military importance in order to develop high-performance AI systems…" (p. 5). Horowitz further notes that those nations that can train and retain skilled engineers will have an edge over others. Nations that have more computing power will also be in a better position to innovate in AI—distant followers can purchase AI systems, but the leaders will innovate, and strength in computing power will make a difference. Cross-sectoral cooperation (in US, public-private cooperation) is crucial for rapid development. The report notes that China's civil-military fusion policy gives the nation a possible advantage.

AI is a form of software (or clusters of software applications) designed and trained to accomplish complicated or complex tasks. Most AI applications and developments are computationally intensive. AI requires advanced computing: combining massive data sets with sufficiently powerful computing to process them. Computer chips are often designed to the specific task of running AI algorithms (especially deep learning algorithms), and here, China may be at a disadvantage, since chipmaking is not an industry mastered by Chinese companies. For these reasons, we include computer science (SC) in the analysis.

This paper looks deeply into these publications to assess the relative capacity of an aspiring China against the leading role of the United States. We expect that late industrializers like China will engage in a learning strategy around emerging technology development, as described by Amsden (1992), observing that learning through imitation rather than slogging through generations of endemic innovation, marks development of late industrializers. Nevertheless, China differs from other late industrializers with a commitment to endemic development. Where learners combine lower wages, state subsidies, and incremental productivity and quality improvements for existing products (Amsden, 1992), China has added active domestic development. We add in here a search for mimetic isomorphism of policy investment and practice, assuming China aims at reaching a point where domestically led innovation, rather than imitation, begins to drive market position. We will test whether China has emerged from a development stage to an independent actor in AI and how the Chinese program compares to that in the United States. Further, we ask whether China can be said to follow the Korean model or whether it constitutes its own model of rapid imitation.

## Grouping research outputs

The methodology involves comparing indicators of China's AI-related output to research output from the US. In order to assess whether China's AI development process follows an isometric pattern, we define seven cohorts of research outputs in the AI and CS fields respectively, based on authorship addresses in scholarly articles curated in DBLP dataset and Microsoft Academic Graph.

(1) Group 1 (Purely China): All the authors of a paper are from China.

(2) Group 2 (Purely USA): All the authors of a paper are from USA.

(3) Group 3 (China): A paper with at least one author from China and no author from the USA.

(4) Group 4 (USA): A paper with at least one author from USA and no author from China.

(5) Group 5 (China & USA): A cooperative work with authors from only and both China and USA

(6) Group 6 (China & non-USA): A cooperative work with authors from China and non-USA countries.

(7) Group 7 (USA & non-China): A cooperative work with authors from USA and non-China countries.

## Tests

To test China's standing in AI against that of the USA, we decompose national capability into five aspects. We then do the tests one by one. In each of these aspects, we expect to see that China will follow (and possibly surpass) the leadership of the United States in AI research. We introduce and explain these important aspects as follows.

(1) Stock of research. This measures the total volume of research output that a country has produced in a scientific field. It tests how China, starting from a low base, compares over time to the US in the research field of AI.

(2) Incremental research. In an evolving timeframe, incremental research depicts the growing trajectories of China and US in the AI field, and examines whether one imitates the first-mover/incumbent in the timeline. This is accompanied by testing whether publications of policy statements lead or lag one another over time to see if the incremental growing of research coincides with policy investment.

(3) Structure of Scientific collaboration. Scientific collaboration is also a vital aspect to evaluate the development of national capability. A reasonable scientific collaboration structure can catalyze scientific innovation. We will test whether China is imitating the United States in choosing its own collaboration partners.

(4) Research quality. Research quality uses indicators of the two countries' performance in the AI field. It may be less difficult for China to catch up in total and incremental volume, but could be much hard to surpass in measures of research quality in a short time. Research quality is operationalized in three proxies: In the first dimension, we use PP-top n% designed for high-quality paper measurement in a nation's level (Wagner et al., 2022); in the second dimension, we use papers published in Top AI conferences and journals; in the third dimension, we use 2-year citations and 5-year citations as a proxy of research quality, respectively.

(5) Research contents. The most direct outcome of mimetic isomorphism is the convergence of what is researched between two countries in the AI field. If China, as the latecomer, produces AI publications that are increasingly similar to the US in content, it can be reasonably inferred that mimetic isomorphism plays a role.

# Measures

In this study, to characterize China's catching-up in CS and AI, we design to consider several main aspects covering research volume, research quality, and research fields. Detailed measures regarding these aspects are introduced below.

**Research volume**: The number of publications and the percentage of the total number of publications are employed to quantify the research quantity.

**Structure of Scientific collaboration:** The Shannon index has been widely used to quantify the diversity in numerous studies, such as discipline diversity (Stirling, 2007), gender diversity (Yang, Tian, Woodruff, Jones, & Uzzi, 2022), et al. In this study, the Shannon index is used to measure the diversity of collaborative countries. The formula is as follows:

$$D_j = -\sum_{i=1}^{n} w_i \ln w_i$$

Where $D_j$ indicates the collaborators' diversity of country $j$, and $w_i$ indicates the proportion of total collaboration count of country $j$ made up of collaborator $i$. The higher the value of $D_j$, the higher the degree of diverse collaborators in country $j$.

**Research quality**: The PP-top1%, the share of high-quality journal/conference papers and citation count are served as proxies for research quality. Specifically, PP-top1% denotes the relative participation in the top-1% highly cited papers of each country (Wagner et al. 2022). We extend the analysis to top 20% as the threshold to obtain the highly cited segment, named PP-top20%. For example, there are 200,000 publications published in 2015, after ranking the publications in descending order by two-year citations, and the top twenty percent of 200,000 is 40,000. Record number 40,000 has 10 citations. Therefore, all publications with greater than or equal to 10 two-year citations belong to the top-20% segment of the DBLP dataset. Assuming China published 80,000 papers in 2015, and 8,000 had 10 or more two-year citations. It is the observed value of China's articles in the top-20% class. The expected value of China's articles in top-20% class is 16,000 (20% of 80,000). Thus, the PP-top20% of China is 0.5 (i.e., the observed value of China's publications in top-20% layer divided by expected value of China's publications in top-20% layer). For the high-quality journal/conference, we select all class A journals/conferences recommended by China Computer Federation (CCF) to represent high-quality journals/conferences in CS[ii]. Moreover, we consider all class A journals/conferences of AI subfield recommended by CCF to represent high-quality journals/conferences in AI. The detailed high-quality journals/conferences lists are shown in Table A1. As for citation count, we use 2-year citation and 5-year citation to measure the research quality respectively. Given that CS is growing rapidly, a 2-year citation window is sufficient for papers from CS discipline to receive citations. It is noteworthy that only citations from conference or journal papers are considered in our study, citations from other types of publications are ignored.

**Research contents:** To test the possible imitation or learning effect between China and USA, we devise an index of topic overlap for the research outputs of the two countries. This index is conceived from three aspects:

(1) the number of all shared keywords in the papers published by purely China (Group 1) and purely USA (Group 2)

(2) the number of shared keywords in the lists of Top 1000 most frequently occurring keywords in Group 1 and Group 2

(3) the Jaccard similarity of the lists of keywords in Group 1 and Group 2:

$$\text{Jaccard similarity} = \frac{\textit{the number of keywords shared by Group 1 and Group 2}}{\textit{total number of keywords in Group 1 and Group 2}}$$

In practice, we used Jaccard similarity instead of the number of overlapping words, because the latter measure is in fact time-dependent when considering time lags between the research contents of USA and China. The reason is that the volume of overlapping words is proportional to both countries' research volume, and the latter in the AI field is almost increasing with time after 2000.

---

[ii] Last Accessed October 25th, 2021, https://www.ccf.org.cn/Academic_Evaluation/By_category/

## Data

The dataset used in this study is the Digital Bibliography & Library Project (DBLP), which is one of the best-curated online bibliography websites for Computer Science (CS). DBLP data is comprised of metadata on CS publications, including authorship information, publication venue, citing relations, and other related information. We downloaded the DBLP released at the beginning of April 2020 and parsed it into a local MySQL database[iii](Tang et al. 2008). There are about 4.8 million publications between 1936-2020. Journal articles and conference papers are considered the major academic output in CS, and we therefore exclude other types of documents (e.g., book, book chapter, etc.). Based on authors' affiliation information, we deployed regular matching to identify the country of each authorship in the paper. 82% of CS publications have at least one identified country. We then retained those publications for which all the authors' country information could be identified. Furthermore, DBLP has been linked to Microsoft Academic Graph (MAG) (Wang et al. 2020), which assigns fields of study (FOS) to each included paper. Following Frank et al (2019), we construct our AI dataset based on papers in the subfields of AI, machine learning, pattern recognition, natural language processing and computer vision. After all the procedures above, we obtain 2,777,378 CS publications and 712,878 AI publications published between 2000 and 2019 for the following quantification of the catching-up process of China.

## Matching

Simply comparing the impact of China-US co-authored and purely China (US) authored papers does not capture the true "impact premium". Thus, we conduct a matching experiment as follows. The China-US co-authored papers are regarded as treatment group, and the control group are selected from purely China (US) authored papers. To make the two groups of papers more comparable, each China-US co-authored paper is matched to one purely China (US) authored paper (without-

---

replacement) with similar basic information. To mitigate interference of inherent differences among papers in different research topics, generations and author composition, we use following criteria for matching:

(1) The purely China (US) authored paper was published in the same year as the corresponding China-US co-authored paper.

(2) The purely China (US) authored paper was published at the same publication venue as the corresponding China-US co-authored paper.

(3) The purely China (US) authored paper has the same number of authors as the corresponding China-US co-authored paper.

The matching process produced three groups of papers: 11,917 China-US co-authored papers, 131,959 purely China authored papers and 136,932 purely US authored papers in CS; 3,229 China-US co-authored papers, 38,269 purely China authored papers and 32,219 purely US authored papers in AI. Thus, each pair has one China-US co-authored paper and one or more control group counterparts. For the pair with more than one control group counterparts, we followed the practice of Farys and Wolbring (2021) to treat this condition: if there are n matches for a China-US co-authored paper, we assign a weight of 1/n for each of these control group counterparts.

*Table 1 Total data and matching sample, 2000-2019*

| Field | Total | Matching | | |
|---|---|---|---|---|
| | | Purely China | Purely US | China-US collaboration |
| CS | 2,777,378 | 131,959 | 136,932 | 11,917 |
| AI | 712,878 | 38,269 | 32,219 | 3,229 |

# Policy Actions and Output

Over five years between 2017 and 2021, both China and the United States increased public spending on AI and greatly increased output of scholarly articles. The Center for Study of Emerging Technologies estimates for 2018 that China and the United States spent about the same amount of funds on AI R&D (Acharya & Arnold, 2019). More recent figures are not available. In 2023, the United States government will invest $1.7 billion in AI R&D.

Table 2 shows the numbers of Web of Science documents published by all nations, by the USA, and China, as well as times cited during those years. The percentage of AI documents cited is higher for the United States is very similar at 73.6 to China's 72.7%, suggesting a level of parity—with both nations' work cited far above the world average of 64%. Category normalized citation impact shows USA leading China and both leading the world. Similarly, the US has a higher percentage of documents in the top 1% most-highly cited works, although China has a greater number of documents in this category. When a search limits national representation, USA has 313 highly cited documents compared to China at 280; joint work is even higher. When combined (China and USA), the number of highly cited documents is 617 out of 28,428 articles and a much higher percentage of documents in the top 1% most-highly cited, showing joint work as higher impact than the work emanating from either nation alone.

*Table 2 Artificial Intelligence articles, China and USA, 2017-2021(Web of Science)*

| Name | # Web of Science Documents | # Times Cited | % Docs Cited | Category Normalized Citation Impact | % Documents in Top 1% MHC | % Documents in Top 10 MHC | # Highly cited |
|---|---|---|---|---|---|---|---|
| Computer Science, Artificial Intelligence – All | 358,547 | 2,984,903 | 63.88 | 1.0 | 0.98 | 3,531 | |
| Computer Science, Artificial Intelligence – USA | 25,583 | 429,691 | 74.51 | 2.1 | 2.52 | 645 | |
| Computer Science, Artificial Intelligence – China | 103,865 | 1,112,125 | 66.76 | 1.3 | 1.52 | 1,579 | |
| Artificial Intelligence – USA(not China) | 67,988 | 1,300,796 | 73.6 | 2.5 | 4.9 | 23.7 | 344 |
| Artificial Intelligence – China (not USA) | 107,112 | 1,519,428 | 72.7 | 1.6 | 3.3 | 17 | 1,109 |
| Artificial Intelligence – USA & China | 28,428 | 697,723 | 83.5 | 2.7 | 6.1 | 28.5 | 617 |

At the policy level, the U.S. government has successively issued a series of policy documents on AI development strategies. During the Obama Administration (2009-2017), the U.S. government released three reports on the development of AI, covering many aspects such as the AI impact on cybersecurity, key areas of AI, and the impact of AI on the U.S. economy. During the Trump Administration (2017-2021), nine AI policies were published, most of which focused on the application side. So far, the Biden administration (2020-2024) has released several reports on AI to develop the comprehensiveness to maintain the advantages of U.S AI development to ensure national security, technological competition, and, particularly, in competition with China. Meanwhile, the U.S. kept increasing the investment into AI. Take NSF as an example: NSF delivered $868 million in AI-related grants establishing several top AI research institutes.

Correspondingly, China, as one of the biggest competitors of the U.S. at this stage in terms of funding, also released many AI-related national-level policies since 2015, such as "Made in China 2025" and "Guiding Opinions on Actively Promoting the "Internet +" Action". The 14th Five-Year Plan for National Economic and Social Development and the Outline of Vision for 2035 of China take the new generation of AI as the primary target area for tackling key scientific and technological frontiers.

Comparing the AI policy between China and the U.S. implies how the two nations invest in policy actions. Judging by the timing of the specific policy issuance, the U.S. and China are nearly identical, with China releasing its first country-level AI policy in 2015, and the U.S. issuing its first state-level AI policy in 2016. However, in terms of content of AI policy, the first AI policy released by China, namely Made in China 2025, focuses on the improvement of high-tech manufacturing. This policy document spends very little space discussing AI technologies. After one year, AI-related action plan was issued by State Council of China, namely the Guidance on Actively Promoting the "Internet Plus" Action. "Internet Plus" regards the Internet and other information technology as a core component of new economic form, which can be integrated with industry, finance, government, even all elements of the economy and society, to spur the growth of innovative industries in China. But the development of AI-technologies is rarely mentioned in this action plan. Regarding United States' AI policies, three AI strategic plans were issued by the U.S. government in 2016, including "Preparing for the Future of Artificial Intelligence"; "The National Artificial Intelligence Research and Development Strategic Plan" (NAIRDSP); and "Artificial Intelligence, Automation, and the Economy". The first two AI strategic plans complement each other. The first AI strategic plan aims to discuss the current status quo of AI development, application fields, and potential public policy issues. The second AI R&D strategic plan specifies the seven strategic directions for the development of AI that the U.S. government gives priority to. The third AI policy pointed out the transformative impact of AI on the U.S. economy in the coming years. To catch up with the western technological prowess in AI, "Next Generation Artificial Intelligence Development Plan" (NGAIDP) was issued by China's State Council in 2017. It is the first time the keyword, "Artificial Intelligence," appeared in the title of a state-led policy, and it is the first official AI policy in China. Comparing NAIRDSP with NGAIDP, these two policies regard the technological research and development as the main task, including developing new AI algorithms, AI systems, ethical use of AI, and the developing of training and testing resources for AI. Moreover, both policies focus on the improvement of AI labor force training and the construction of AI technology standards. As for the resource allocation, policies in both countries call for accelerating advances in AI through collaboration. Certainly, differences between the two policies exist. The NAIRDSP aims to pay more attention to the micro field, while the goal of NGAIDP is to create a comprehensive layout of industrial chain from the macro perspective. Another notable action policy was released by Ministry of Industry and Information Technology at the same year, namely "The Three-Year Action Plan to Promote the Development of a New Generation of Artificial Intelligence Industry". The plan highlighted the key research and development priorities in AI, indicating China's lag in software and hardware foundations behind those of foreign developed countries.

During the Trump Administration, nine AI-related policies were published, most of which focused on the application side. For example, "Maintaining American Leadership in Artificial Intelligence" is released by Executive Office of the President Donald Trump in 2019. This presidential document emphasized the protection of the American AI technology and development of AI standards, and the training of AI workforce is of concern to this policy document. As for China, state-lead AI policies focus on the AI technological standard, governance norms and education after 2017, so highly similar, with documents such as "AI Innovation Action Plan for Institutions of Higher Education" (2019), "The Ethical Norms for the New Generation Artificial Intelligence (2021)", "Guidelines for the Construction of a National New Generation Artificial Intelligence Standards System (2021)". In addition, China released fewer country-level AI policies than the U.S. All in all, we believed that China's AI policymaking appears to be lagging behind and imitating that of the US, and these phenomena support mimetic isomorphism.

# Results of AI research output analysis

## Dynamic trends and overturned disparity

From a dynamic aspect of research productivity, we find an emerging reversed disparity between China and the United States (Figure 1). In the first decade of the 21st century, as expected, the United States (both USA and purely USA) produced the largest number of yearly publications. China (both China and purely China) produces fewer articles than the USA; the United States' publication growth entered a plateau starting from around 2011 in both research fields, while China's productivity kept a strong upward trend, especially in the AI related research. In fact, a reversed trend is emerging in the gap between the two countries' yearly research outputs. For the participative research every year in the CS field, China caught up to the United States in 2018; for the solely undertaken research, the catching up occurred even one year earlier (in 2017).

In AI, China outproduced the United Stated in 2014 in China-only research and surpassed the USA in 2015 in collaborative research, shown in Figure 1, which also shows the growth rates for CS and AI research. Since 2015, China continued to out-produce the United States. We expect China to continue to outpace the growth of the United States as suggested by the trend lines in Figure 1 The tendency becomes clear that China is catching up to the United States in yearly amount of AI papers shown in Table 2 (above).

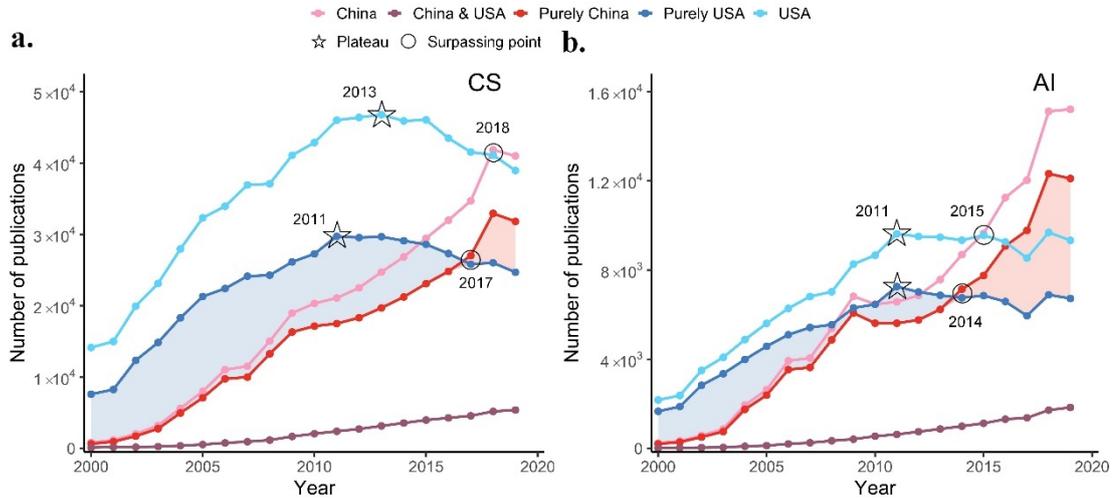

*Figure 1 Yearly growth of research outputs, with the gap highlighted between purely USA and purely China*

Replacing publication counts with relative share in a research field, the disparity becomes more prominent. The United States' share shows an overall downward trend. This suggests that China's research in the AI field (as well as in the broad CS field) is not only growing itself but is increasing its percentage share as it brings along the growth of the whole field. Such a growth of Chinese AI research reflects the policy measures implemented by the Chinese government in recent years for fostering the development of AI. It should be noted that the time points of production growth should not be tied directly to the policy statements, as production will lag policy investment by quite a few years.

The key time-points regarding China's AI related policies are marked in Figure 2. As early as in 2012, the Ministry of Science and Technology released a twelfth five-year plan for intelligent manufacturing technology development[iv]. Next in 2015, the State Council successively issued two policies – Made in China 2025[v] and the Guidance

---

on Actively Promoting the "Internet Plus" Action[vi], to invest heavily in China's own innovations and to reduce its reliance on foreign technologies. It was also in 2015 that China overtook the United States in the amount of AI publications. One year later, in 2016, the central ministries and commissions issued two related development plans for the robotics [vii] and Internet [viii] industries, to accelerate the application of Artificial Intelligence technologies in various fields and promote the level of intelligent services in China. Furthermore, in 2017, a development plan[ix] and an action plan[x] specific to a new generation of Artificial Intelligence were issued respectively by China's State Council and Ministry of Industry and Information Technology, aiming to make the nation one of the world leaders in AI in the future. These points in time regarding China's AI related policies were accompanied by China's AI publication growth. In terms of AI policy, China's first state-led policy targeting AI was released in 2017, one year after the U.S. AI policy was issued (The National Artificial Intelligence Research and Development Strategic Plan (2016)). China began to catch up with the USA after 2000 and then surpassed it in 2014, which supports our hypothesis that China is rapidly catching up with the USA, at least in quality output. Additionally, the United States' research in the AI field stopped declining after 2017, and began increasing its percentage share, which may also reflect policy attention. The effect of AI policy on promoting AI publication is not a direct relationship but indicates underlying support. China's artificial intelligence policy statements lag behind that of the U.S., although it is difficult to say that direct imitation is at work.

---

[vi] The "Internet Plus" Action(关于积极推进"互联网＋"行动的指导意见)
[vii] Development plan for the robotics industry (2016-2020)(机器人产业发展规划（2016-2020）)
[viii] Three-Year Guidance for Internet Plus Artificial Intelligence Plan (2016-2018) ("互联网+"人工智能三年行动实施方案(2016-2018))
[ix] Development Plan of the New Generation Artificial Intelligence (新一代人工智能发展规划).
[x] Three-Year (2018-2020) Action Plan for Promoting Development of a New Generation Artificial Intelligence Industry(促进新一代人工智能产业发展三年行动计划(2018-2020 年))

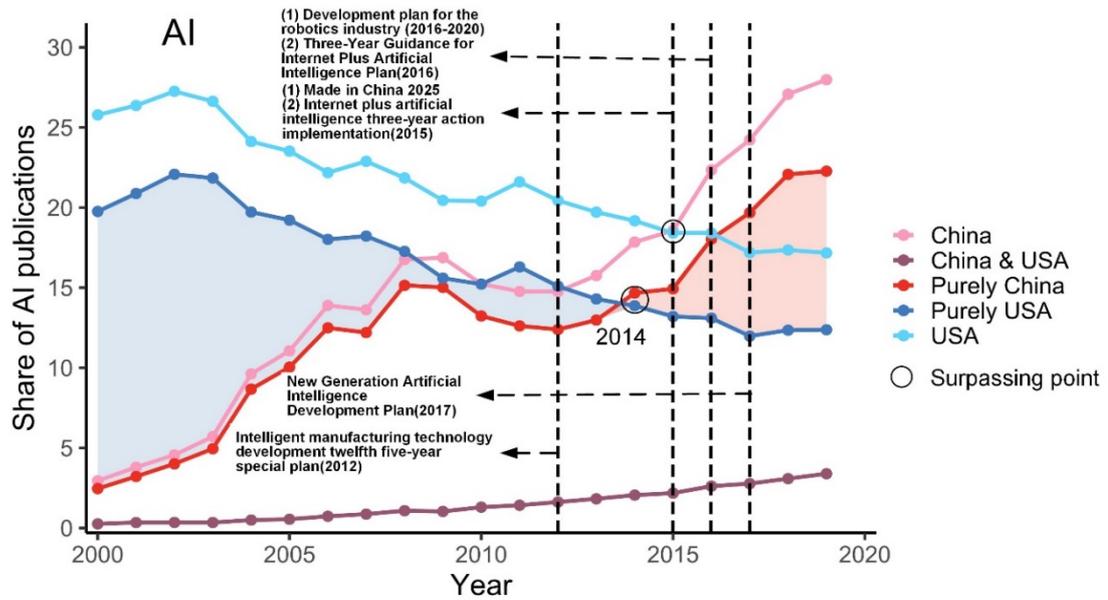

*Figure 2 AI Publication Share and China's AI related policies*

## Reduced disparity and similar research structure

We compared three fundamental aspects of research output in China and the United States: total research output, total citation impact, and citation sources. We anticipated that the two countries would share similar conditions due to China's rapid imitation.

Figure 3(a-b) demonstrates, consistent with other findings (Lundvall & Rikap, 2022), that the United States has more academic publications in CS and AI-related fields than China. According to the data, and as expected, the two nations' scientific and technical knowledge production is becoming increasingly similar. In the field of AI, the gap between the United States and China has significantly narrowed over time, with the United States possessing only 17,760 (14%) more AI publications than China. This difference decreases to 2,630 (2.5%) publications when we consider only AI research conducted by each nation (the purely China bar and purely USA bar). This indicates that China's catching-up strategy is bearing fruit, as China is catching up to the United States in terms of AI research outputs.

In terms of total citation impact, data indicates that the two nations have comparable standings. However, further investigation suggests two more profound observations. The first observation demonstrates that the United States has a greater overall impact than China, both in the AI and broader CS fields. To test this hypothesis in a statistically sound manner, we regard China-US co-authored papers as the treatment group and construct two control groups (see details in the Data and Method section) consisting of purely China papers and purely US papers (see details in the Data and Method section). Figure 3 (c-d, e-f) demonstrates that the Kruskal-Wallis test reveals a significant difference between the three pairs of papers with different authors.

China and the United States receive the highest proportion of citations from their own country, but they also occupy a sizeable portion of each other's citation pool (Figure 4). In addition, Europe, South Korea, and Japan are all significant citation sources for the two nations. This shows a somewhat similar structure of citation sources for the two nations. Yet Chinese publications receive a significantly lower proportion of international citations than American publications; this may be an audience effect rather than a measure of quality (Wagner et al., 2019). Chinese AI research receives citations from countries other than China at a rate of 50 percent or less, whereas US AI research receives citations from countries other than the United States at a rate greater than 50%.

Similarities in research volume, citation impact, and citation sources support our hypothesis that the fundamentals of AI research in the two countries are nearly identical at this writing.

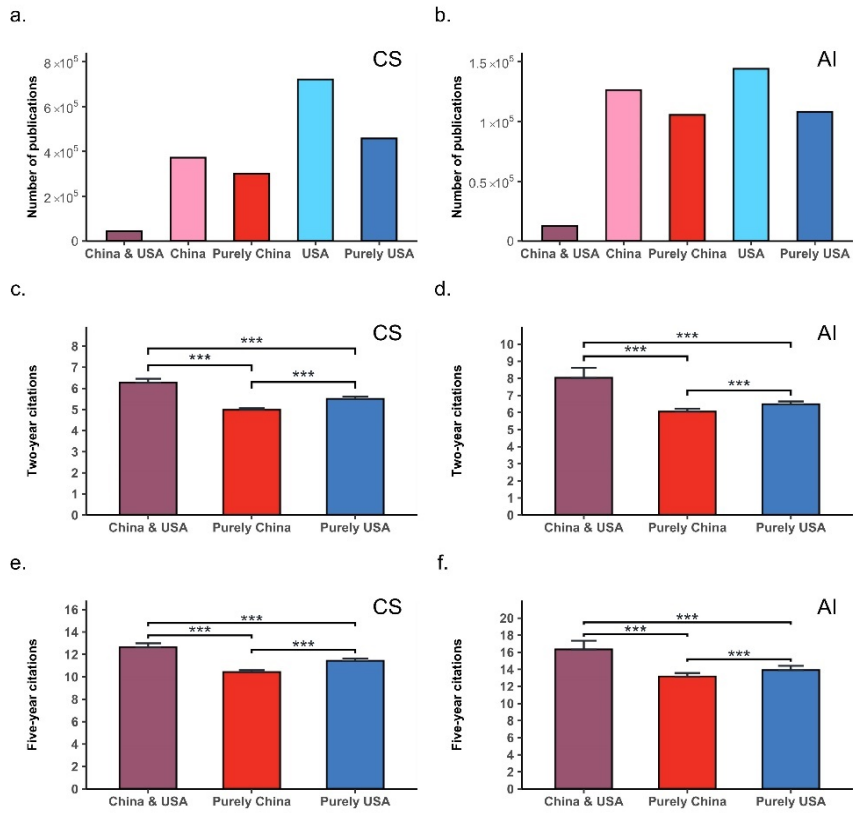

*Figure 3 Research output (a-b) and citation impact (two-year, c-d; five-year impact, e-f)*

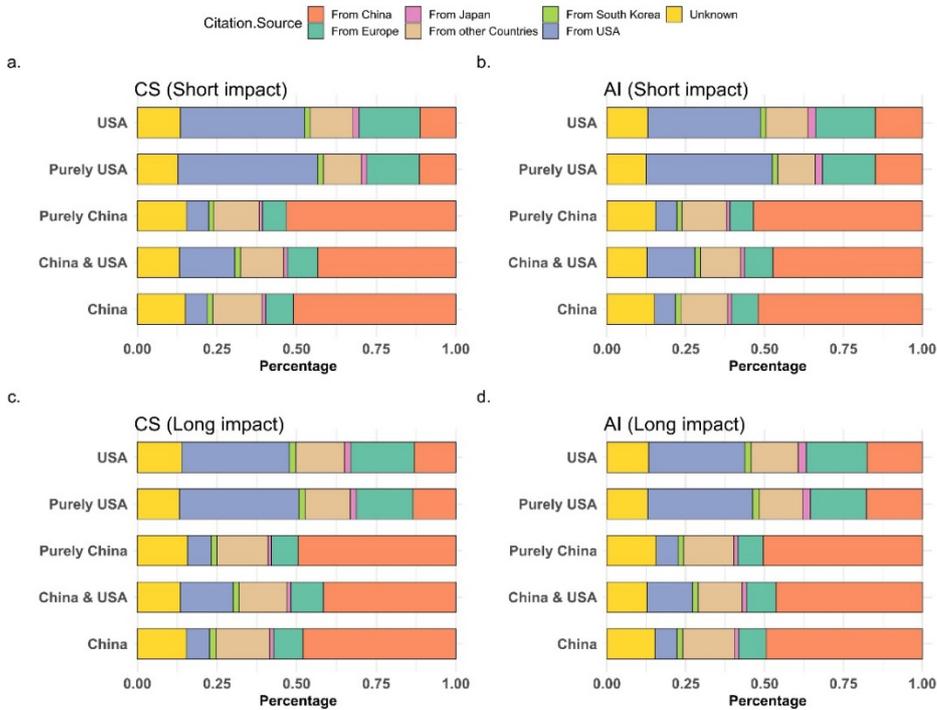

*Figure 4 Sources of citation through the timeframe of 2000-2019 (short term= 2 years, long term = 5 years)*

# China's independent growth and simplex collaboration

The coincident growth of publications by China and China-US collaboration depicted in Figure 1 prompts further inquiry: what are the sources of Chinese AI research growth? Does the growth stem primarily from the benefits of external collaborations (particularly with the United States), China's own commitment to AI research, or other factors? We sample a subset of AI publications with at least one Chinese author and then plot the proportions of the principal forms of authorship in Figure 5. More than 70% of AI papers were written independently by Chinese authors. China-U.S. collaborative papers account for less than 12%. In contrast, China's non-US collaborations, including those with European nations, have always outnumbered those with the United States. After 2008, the gap in Figure 5 regarding China's collaboration preference (non-U.S. collaborations versus U.S. collaborations) widens. This suggests, as anticipated, that China's growth reflects independent commitments and efforts in AI research, with external learning playing a secondary role. Significantly, China plays a dominant role in these collaborative relationships, as Chinese first-authored papers have accounted for more than half of all collaborative papers since 2000, and this percentage has increased to over 75% in recent years (after 2018) for US collaboration and European collaboration (Figure 6a). The data in Figure 6a also suggests that, in the field of AI, Chinese authors lead research collaborations with American partners more frequently than with European or other non-American researchers. In recent years, Chinese authors have occupied between 40% and 45% of the last-author position, which can denote the corresponding authors who leads research projects (Figure 6b) (Lui & Fang, 2014).

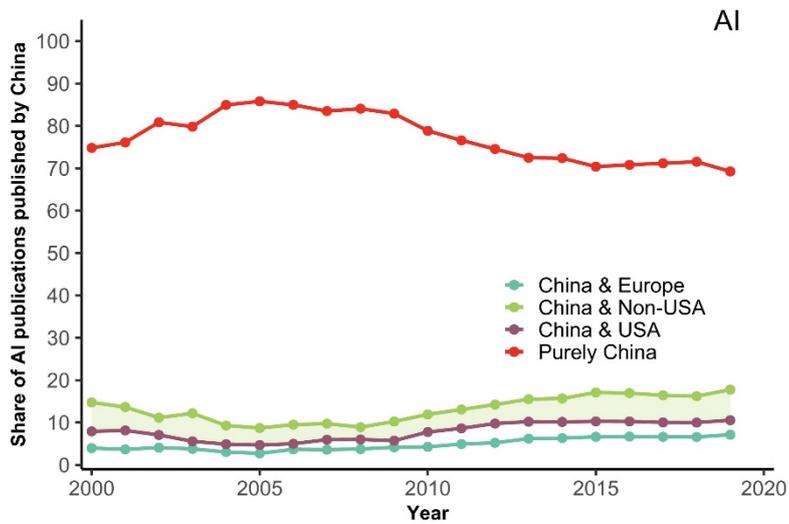

*Figure 5 China's independent growth in collaborative AI research*

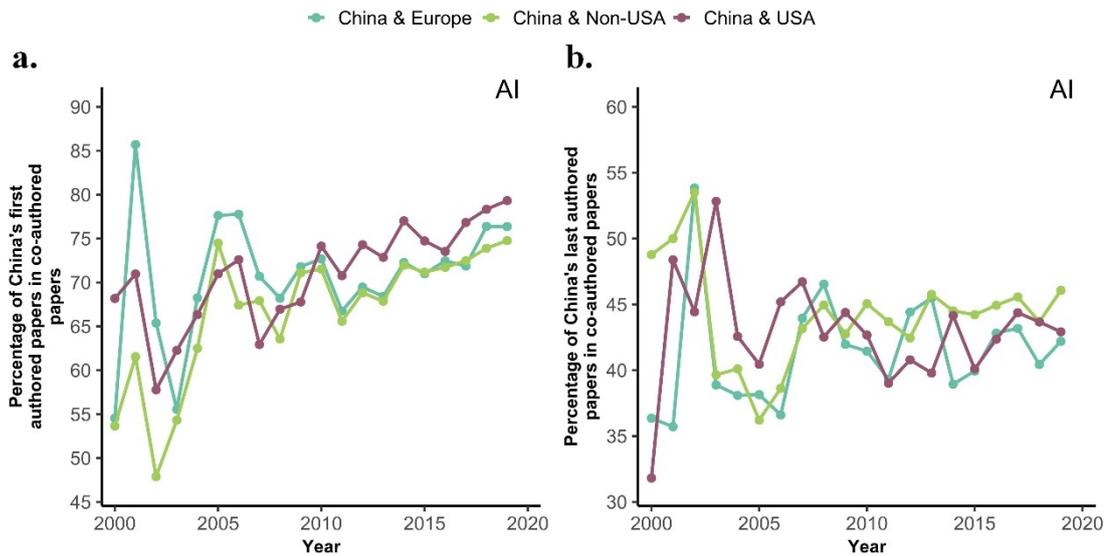

*Figure 6 China's dominant role in collaborative research in the AI field*

We conducted additional tests to compare the diversity of the fundamental knowledge structure in China and the United States. A diverse economy is associated with innovation. Shannon diversity is a measure indicating the diversity of countries, with a higher value indicating greater diversity and less reliance on a few collaborators. The convergence of the two nations in terms of complexity was anticipated to illustrate mimetic isomorphism. Figure 7 demonstrates that the two are indeed converging, particularly in AI research, supporting our isomorphism hypothesis. While the United States maintains a more diverse structure for collaboration, China is gradually increasing its diversity.

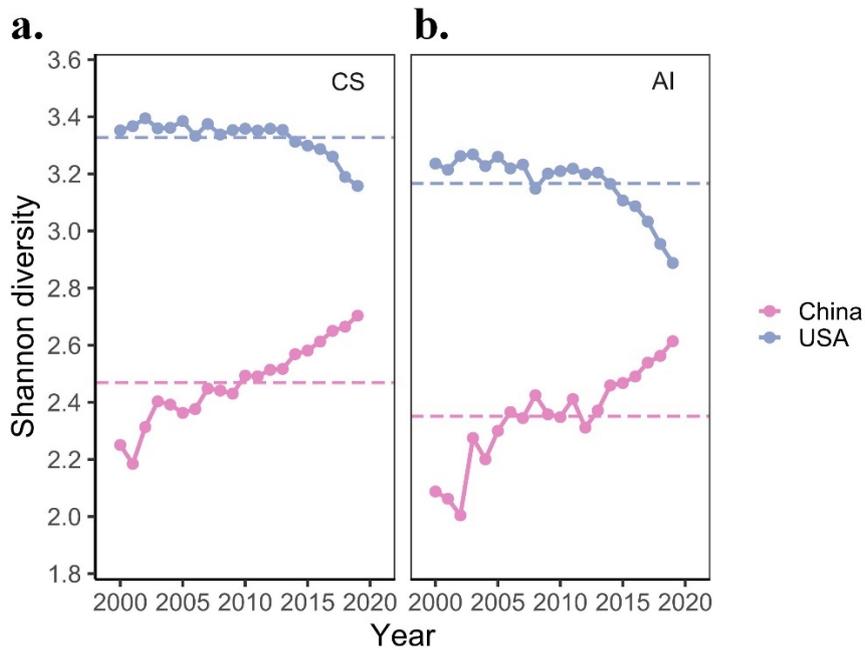

*Figure 7 Collaboration diversity of China and the U.S.*

Figure 8 vividly depicts the relatively complex structure of the United States in AI research compared to the relatively simple structure of China. The United States' most frequent collaborators have shifted from the United Kingdom to China over the past two decades, while China's top collaborator has always been the United States. Singapore, Japan, Australia, the United Kingdom, Canada, France, and Germany are among China's top 10 collaborators, which have remained relatively stable. The United States continues to collaborate with numerous nations, the list of which is constantly evolving. Japan, for instance, once ranked second but dropped out of the top ten in 2017; India, which previously ranked outside the top ten, has recently risen to become the United States' fourth-ranked collaborator. Figure 8 demonstrates that, compared to the United States, China's collaboration structure is relatively simple, but growing diversity (Figure 7).

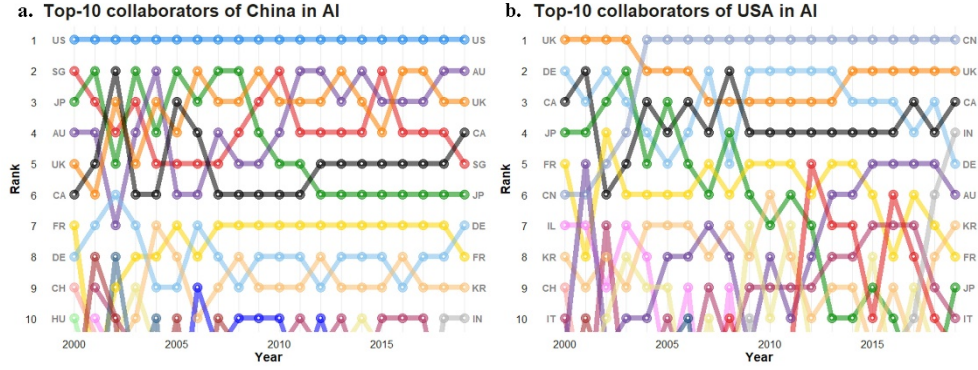

*Figure 8 The evolution of (a) Top 10 collaborative partners of China and (b) the United States in the AI field since 2000*

## Large gap in research quality

The quality of AI research in China and the United States was evaluated. We anticipated that mimetic isomorphism would be manifested by a convergence of the two countries in terms of research quality, which is proxied by papers with a high impact factor and publication in prestigious academic venues.

As illustrated in the section on measures, we use PP-top20%, an indicator similar to the measure of PP-top1% proposed by Wagner et al. (2022), to evaluate the research quality of each type of authorship. A PP-top20% value of 1 indicates that a country has exactly the expected number of highly cited papers.

Figure 9 depicts a gap between the United States and China in their PP-top20% values, but one that is narrowing. Our hypothesis that the two nations' research quality is nearing parity is supported by the fact that the two nations' high-impact paper outputs are converging. United States AI research is unquestionably of higher quality. In particular, its PP-top20% value in the AI field (Figure 9b) is almost always greater than 1.5 and occasionally reaches twice the expected value. In contrast, China began with a value of approximately 0.5, which was only half of the expected value, and increased to near 1 while still trailing the United States. A similar but more prominent situation exists in the CS field (Figure 9a), with China's recent research in this field appearing to be of higher quality (PP-top20% value) than the more specific subfield of AI. An intriguing finding is that despite China's mediocre performance in quality of research by itself, its collaborative research is of high quality and has strong growth momentum, regardless of whether the collaboration is from the United States or other countries (the China & USA curve) (the China & Non-USA curve).

However, the United States' research collaborations with other nations (the USA & Non-China curve) do not yield as high impact scores as its collaborations with China (the China & USA curve). This suggests that the United States also benefits from collaboration with China, and that these benefits are even greater than those derived from collaboration with other nations.

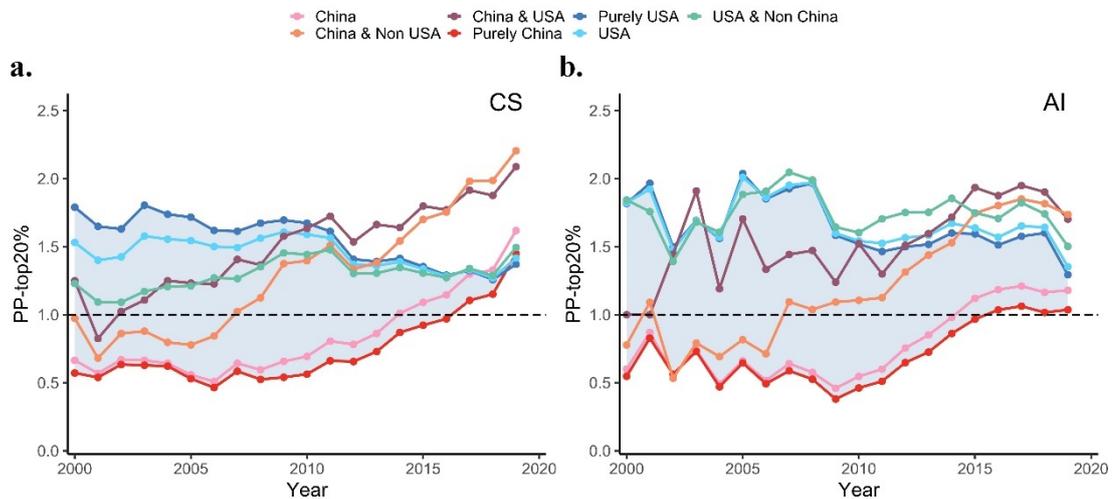

*Figure 9 Variation in the PP-top20% value as a measure of the quality of research*

Figure 10 indicates that China lags significantly behind the United States in terms of the proportion of publications in leading AI conferences and journals. However, it demonstrates that China and the United States are also approaching one another, supporting our hypothesis that research quality is converging. In terms of top AI conference publications, the United States holds a substantial advantage over China (Figure 10d) in terms of both total volume and annual output. China publishes a growing proportion of papers at leading AI conferences, but the United States (the shaded area) remains far ahead. A similar situation exists in the field of CS (Figure 10c). In both AI and CS, conferences are the primary medium for academic publishing, and top conferences are the preferred venue for leading researchers to publish and disseminate their high-quality research.

The United States' advantage over China in leading AI journals is also evident (Figure 10b), although this disparity is rapidly narrowing. The share of U.S. papers is gradually declining, as China's share slowly increases each year. China lags significantly behind the United States in both collaborative (the China curve) and independent (the Purely China curve) research. A language issue for Chinese authors and presenters may be at work here.

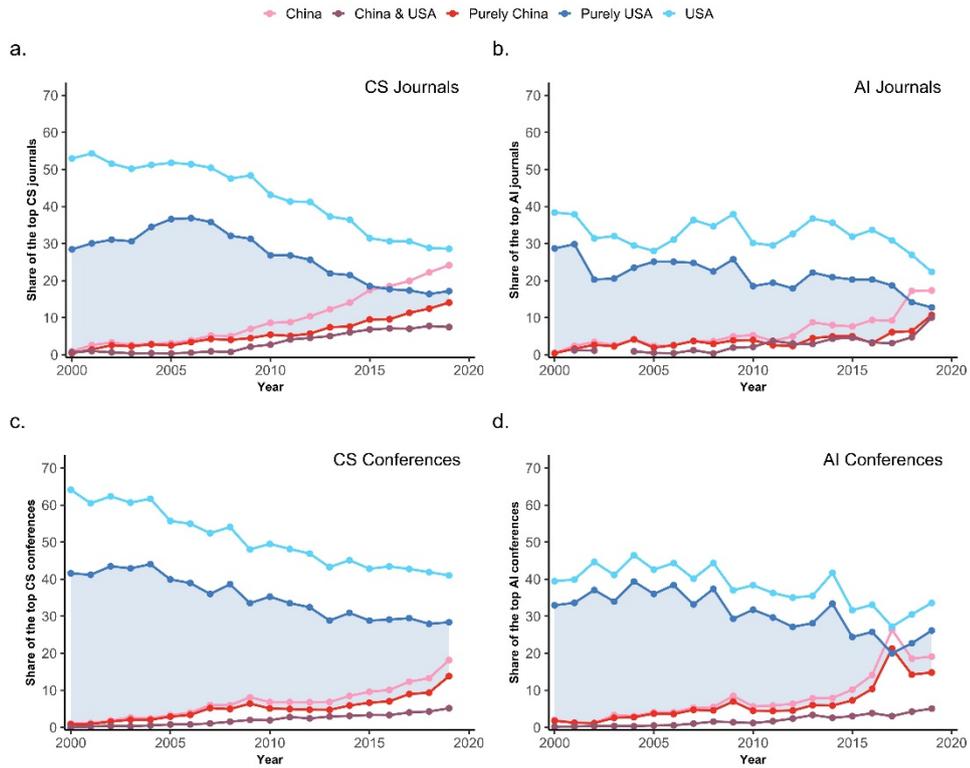

*Figure 10 Share in top publication avenues in the AI and CS fields*

## Convergence and overlap of research topics

Finally, we examined the overlap of AI research topics between China and the United States. Research topics reflect the scope and velocity of research. We anticipated that mimetic isomorphism would be demonstrated by a time lag between the leader and the follower, as well as the convergence of their research pace.

We started with three scenarios: (1) China and the United States conduct AI research at the same rate; (2) the United States lags behind China in AI research; (3)

China lags behind the United States in AI research, but we only present scenario 3 since 1 and 2 were not shown. In the third scenario, there is a high probability that mimetic isomorphism will occur, and that China will learn from imitating the United States.

In fact, China's AI research profile is comparable to that of the United States in previous eras, although the gap between 2000 and 2019 has shrunk. This confirms Scenario 3, that China lags behind the United States in terms of scope of research topics, with similarities increasing and gaps narrowing over time. Using the US publication date as the reference time (Figure 11a) and basing our calculations on the greatest similarities, we determined the exact time lags in each year. From 2000 to 2009, the time lag between China and the United States kept an average value of 5.6 years, with the largest time lag of 9 years occurring in 2007 (which means the AI research China did in 2016 has been done by the US in 2007). Eventually, however, the time lag shrinks from five years in 2010 to one year in 2015, and it disappears in 2018. The results presented here indicate that there is, in fact, a systematic delay in synchronization between AI research topics in China and the United States, and that this divergence in time has been decreasing over the past few years, supporting our hypothesis of isomorphism on the part of China in following the United States—a gap that appears to have closed as shown in Figure 11.

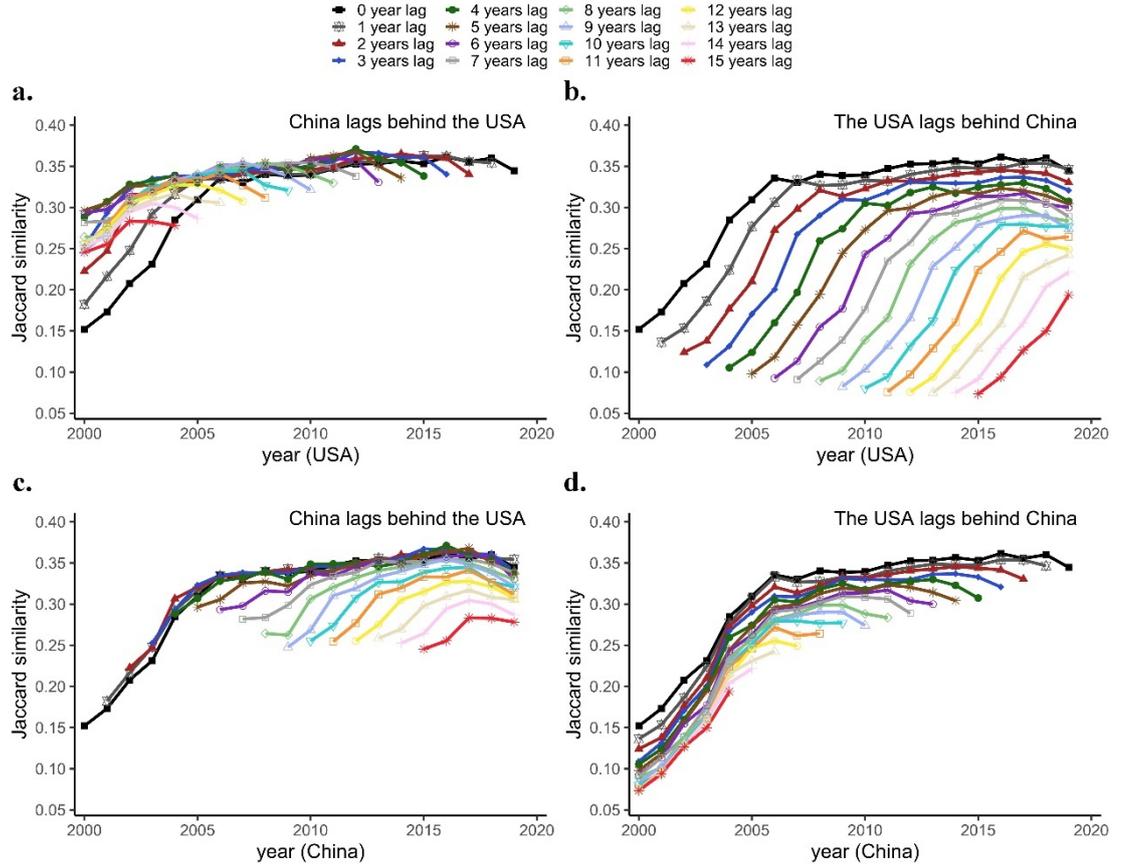

***Figure 11 Lead-Lag Between Frontier Research in China and the United States.*** *Figure 11 presents similarity of AI research topics between China and the USA year by year, with varying assumptive time lags: (a) and (c) assumes China lags behind the USA, where (a) sets the USA publication time as the X axis, and (c) sets China publication time as the X axis. (b) and (d) assumes the USA lags behind China, where (b) sets the USA publication time as the X axis, and (d) sets China publication time as the X axis. Interval = lag time.*

# Discussion

The metrics have limitations, but our analysis shows similar AI research structure between China and the United States, suggesting a mimetic isomorphic process that China has undergone to grow a very substantial AI research sector. In just 20 years, China has closed a vast chasm of capability with the United States. Steered by policy, investment, engagement, and imitation, China is rapidly catching up to the United States in AI research through learning (including sending people abroad and attracting them back to China[xi], imitation, and internal investment). Since 2014, Chinese researchers have published more papers annually than U.S .researchers. The quality measure shows that China continues to lag the U.S. in at least one measure, but in the measure of top 1% most highly cited, China is making an excellent showing. At the current rate of improvement, all things being equal, in the top 20% most highly cited, China will overtake the United States in AI research in years to come. China's diversity does not have the complexity of the U.S., but China's diversity is also being enriched. China lagged the U.S. in researching frontier topics, but this gap appears to have closed, and China is researching at the frontiers.

China's emergence in the field coincides with several propitious developments in the research environment that have favored China's rapid rise. It has become the norm for researchers to share their most recent advancements on open sharing sites such as GitHub. This dissemination practice increases the opportunity for latecomers to quickly close the knowledge gap with the early adopters (Li, Tong & Xiao, 2021). Moreover, China's large user base, generating massive amounts of data resulting in an abundance of material, can be tapped for AI applications. China's regulatory structures around data use are comparatively lax, providing China with advantages in the implementation of AI applications and the advancement of related research. Moreover, the Chinese government's role in establishing its national innovation system for AI development has been crucial (Lundvall & Rikap, 2022). In 2010s and 2020s, China released a series of AI development-promoting policies with accompanying policy environment conducive to AI's rapid expansion.

When it comes to translation into products as measured by patents, China lags the United States and the European Union, according to Castro et al. (2019). Not only does China own fewer high-impact patents despite its large number of total patents, but the majority of its patents are issued only by the Chinese patent office, as opposed to foreign

---

[xi] In 2019, a report by Nikkei Asian Review stated that China had attracted over 1,600 AI experts and researchers back to China from the United States and other countries through various talent recruitment programs, such as the Thousand Talents Plan.

offices such as the USPTO, EPO, and WIPO, suggesting a lower diffusion rate and less competitiveness worldwide. In addition, unlike the United States, the majority of patent owners in China are non-profit organizations such as universities, institutes, and government agencies (China Institute for Science and Technology Policy at Tsinghua University, 2018) as opposed to corporations in the USA. China's lag in chip technology is a fundamental weakness, given the importance of that technology to AI development.

According to a report by Stanford University (Zhang et al., 2021), China is ranked ninth in the Global Vibrancy Ranking of AI, taking into account a variety of economic metrics such as skill penetration, AI hiring index, AI private investment, and number of AI firms. The ability of a nation to cultivate AI companies is crucial to its competitiveness. These companies provide organizations and individuals with the resources and services necessary to adopt AI in their operations. In 2020, only 398 Chinese AI companies received funding over $1 million. This is less than half the number of European AI firms (890) and a quarter of American AI firms (2130) that received the same amount of funding (Castro & McLaughlin, 2021). According to research (Zhang et al., 2021; Lundvall & Rikap, 2022; Castro & McLaughlin, 2021), China-based AI start-ups received $9.9 billion in private equity and venture capital in 2020, significantly less than the $23.6 billion received by AI start-ups in the United States. China must also devote a great deal of effort to the development of global AI leaders in order to advance its economy.

It appears that China's policy and practice has involved active mimetic isomorphism in topics, training, and institution building. International engagement, especially training of students abroad, has been a lynchpin in China's strategy. The imitation of US output appears to have peaked in 2018; China may be moving into an independent phase of AI development. FDI has played a part in China's growth, but to a lesser extent than has been reported of South Korea when it was growing its information technology industries. These observations suggest that China's model of development differs from that of other Asian Tigers or earlier of Japan's export-led policy of the 1970s.

Among potential future studies, institutional-level analyses of AI research would add to the literature. Existing studies have observed an increase in the number of institution-level collaborations in this domain (i.e., inter-institution collaborations, Shao, Yuan, & Wang, 2020). These and other authors find an increasingly central role of industrial contributions to AI, particularly in the United States (Frank, Wang, Cebrain, & Rahwan, 2019; Ahmed, Wahed & Thompson, 2023). Industrial contributions can be more complicated to study because they are less likely to publish. Investigating the various roles of academia, industry, and academia-industry collaborations, as well as the influence of these works, would have significant implications for the future AI policies, workforce training, regulations, and developments of major nations.

# Conclusion

In this study, we sought to characterize China's catching-up process from the mimetic isomorphism angle. We test whether the processes of China's AI development follow an isometric pattern based on five dimensions, including stock of research, incremental research, structure of scientific collaboration, research quality and research content. For the stock of research, the U.S. still publishes more AI and CS publications than China in stocks, but the gap between the two countries is gradually narrowing over time, with China's annual publication count even surpassing the U.S.'s in AI in 2014, and we see isomorphism in this dimension. Moreover, China's AI policy lags behind that of the US, but China's AI publications are increasing over time, so we cannot conclude that policy investment is driving the incremental growth of research. The results thus do not support hypothesis (2). Regarding structure of scientific collaboration, the diversity measure of China and US is getting close over time, upholding the hypothesis (3). In terms of research quality, there is still a certain gap between the United States and China in their PP-top20% values and the proportion of publications in top venues, but China's attempt to close the gap is working, and it supports our hypothesis of isomorphism in research quality. Finally, we examined the overlap of AI research topics between China and the United States. The results suggest that the research scope of China and the United States in AI fields overlaps with a time lag, strongly upholds the hypothesis (5).

Appendix A

Table A1 The high-quality journals/conferences

| No. | Subfield | Type | Venue |
|---|---|---|---|
| 1 | Computer Architecture/ Parallel and Distributed Computing/Storage System | Journal | ACM Transactions on Computer Systems |
| 2 | | | ACM Transactions on Storage |
| 3 | | | IEEE Transactions on Computer-Aided Design of Integrated Circuits and System |
| 4 | | | IEEE Transactions on Computers |
| 5 | | | IEEE Transactions on Parallel and Distributed Systems |
| 6 | | Conference | ACM SIGPLAN Symposium on Principles & Practice of Parallel Programming |
| 7 | | | Conference on File and Storage Technologies |
| 8 | | | Design Automation Conference |
| 9 | | | High Performance Computer Architecture |
| 10 | | | IEEE/ACM International Symposium on Microarchitecture |
| 11 | | | International Conference for High Performance Computing, Networking, Storage, and Analysis |
| 12 | | | International Conference on Architectural Support for Programming Languages and Operating Systems |
| 13 | | | International Symposium on Computer Architecture |
| 14 | | | USENIX Annul Technical Conference |
| 15 | Computer Network | Journal | IEEE Journal of Selected Areas in Communications |
| 16 | | | IEEE Transactions on Mobile Computing |
| 17 | | | IEEE/ACM Transactions on Networking |
| 18 | | Conference | ACM International Conference on Applications, Technologies, Architectures, and Protocols for Computer Communication |
| 19 | | | ACM International Conference on Mobile Computing and Networking |
| 20 | | | IEEE International Conference on |

| | | | Computer Communications |
|---|---|---|---|
| 21 | | | Symposium on Network System Design and Implementation |
| 22 | | Journal | IEEE Transactions on Dependable and Secure Computing |
| 23 | | | IEEE Transactions on Information Forensics and Security |
| 24 | | | Journal of Cryptology |
| 25 | Network and Information Security | Conference | ACM Conference on Computer and Communications Security |
| 26 | | | European Cryptology Conference |
| 27 | | | IEEE Symposium on Security and Privacy |
| 28 | | | International Cryptology Conference |
| 29 | | | Usenix Security Symposium |
| 30 | | Journal | ACM Transactions on Programming Languages & Systems |
| 31 | | | ACM Transactions on Software Engineering and Methodology |
| 32 | | | IEEE Transactions on Software Engineering |
| 33 | | Conference | ACM SIGPLAN Conference on Programming Language Design & Implementation |
| 34 | | | ACM SIGPLAN-SIGACT Symposium on Principles of Programming Languages |
| 35 | Software Engineering/System Software/ Programming Language | | ACM SIGSOFT Symposium on the Foundation of Software Engineering/ European Software Engineering Conference |
| 36 | | | ACM Symposium on Operating Systems Principles |
| 37 | | | Conference on Object-Oriented Programming Systems, Languages, and Applications |
| 38 | | | International Conference on Automated Software Engineering |
| 39 | | | International Conference on Software Engineering |
| 40 | | | International Symposium on Software Testing and Analysis |
| 41 | | | USENIX Symposium on Operating Systems Design and Implementations |
| 42 | Database/Data Mining/Content | Journal | ACM Transactions on Database Systems |

| 43 | Retrieval | | ACM Transactions on Information Systems |
|----|-----------|------------|------------------------------------------|
| 44 | | | IEEE Transactions on Knowledge and Data Engineering |
| 45 | | | The VLDB Journal |
| 46 | | Conference | ACM Conference on Management of Data |
| 47 | | | ACM Knowledge Discovery and Data Mining |
| 48 | | | IEEE International Conference on Data Engineering |
| 49 | | | International Conference on Research on Development in Information Retrieval |
| 50 | | | International Conference on Very Large Data Bases |
| 51 | Computer Science Theory | Journal | IEEE Transactions on Information Theory |
| 52 | | | Information and Computation |
| 53 | | | SIAM Journal on Computing |
| 54 | | Conference | ACM Symposium on the Theory of Computing |
| 55 | | | ACM-SIAM Symposium on Discrete Algorithms |
| 56 | | | Computer Aided Verification |
| 57 | | | IEEE Annual Symposium on Foundations of Computer Science |
| 58 | | | IEEE Symposium on Logic in Computer Science |
| 59 | Computer Graphics and Multimedia | Journal | ACM Transactions on Graphics |
| 60 | | | IEEE Transactions on Image Processing |
| 61 | | | IEEE Transactions on Visualization and Computer Graphics |
| 62 | | Conference | ACM International Conference on Multimedia |
| 63 | | | ACM SIGGRAPH Annual Conference |
| 64 | | | IEEE Virtual Reality |
| 65 | | | IEEE Visualization Conference |
| 66 | Human-computer Interaction and Pervasive Computing | Journal | ACM Transactions on Computer-Human Interaction |
| 67 | | | International Journal of Human Computer Studies |
| 68 | | Conference | ACM Conference on Computer Supported Cooperative Work and Social |
| | | | Computing |
| 69 | | | ACM Conference on Human Factors in Computing Systems |
| 70 | | | ACM International Conference on Ubiquitous Computing |
| 71 | Artificial Intelligence | Journal | Artificial Intelligence |
| 72 | | | IEEE Trans on Pattern Analysis and Machine Intelligence |
| 73 | | | International Journal of Computer Vision |
| 74 | | | Journal of Machine Learning Research |
| 75 | | Conference | AAAI Conference on Artificial Intelligence |
| 76 | | | Annual Conference on Neural Information Processing Systems |
| 77 | | | Annual Meeting of the Association for Computational Linguistics |
| 78 | | | IEEE Conference on Computer Vision and Pattern Recognition |
| 79 | | | International Conference on Computer Vision |
| 80 | | | International Conference on Machine Learning |
| 81 | | | International Joint Conference on Artificial Intelligence |